\def\year{2020}\relax
\documentclass[letterpaper]{article} 
\usepackage{hyperref}
\usepackage{aaai20}  
\usepackage{times}  
\usepackage{helvet} 
\usepackage{courier}  
\usepackage[dvipsnames]{xcolor}
\usepackage{graphicx} 
\usepackage{graphicx}  
\usepackage{amssymb}
\usepackage{amsmath}
\usepackage{pifont}
\usepackage{multirow}
\usepackage{booktabs}
\usepackage{enumitem}

\renewcommand{\url}[1]{\texttt{\textcolor{black}{#1}}}

\newcommand{\gcheck}{\textcolor{ForestGreen}{\ding{52}}}
\newcommand{\rex}{\textcolor{BrickRed}{\ding{56}}}

\newcommand{\hlc}[2][yellow]{{%
    \colorlet{foo}{#1}%
    \sethlcolor{foo}\hl{#2}}%
}

\newcommand{\correctans}[1]{\hlc[cyan!30]{#1}}
\newcommand{\incans}[1]{\hlc[pink!80]{#1}}

\newcommand{\dataset}{\textbf{PIQA}}
\newcommand{\datasetlong}{\textbf{Physical Interaction: Question Answering}}
\newcommand{\pig}{\raisebox{-0.3pt}{\includegraphics[width=11px]{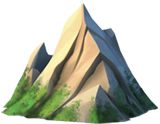}}}

\newcommand{\goal}{\textbf\textsc{\textcolor{YellowOrange}{\scriptsize{\textsc [Goal]}}}}
\newcommand{\sol}[1]{\textbf\textsc{\textcolor{NavyBlue}{\scriptsize{[Sol#1]}}}}

\usepackage{xcolor}
\usepackage{soul}

\title{PIQA: Reasoning about Physical Commonsense in Natural Language}

\author{Yonatan Bisk$^{1,2,3,4}$ \hspace{1.25em} Rowan Zellers$^{1,4}$ \hspace{1.25em} Ronan Le Bras$^1$ \hspace{1.25em} Jianfeng Gao$^2$ \hspace{1.25em} Yejin Choi$^{1,4}$\\
$^1$Allen Institute for Artificial Intelligence\hspace{1.5em}
$^2$Microsoft Research AI\hspace{1.5em}
$^3$Carnegie Mellon University\\
$^4$Paul G. Allen School for Computer Science and Engineering, University of Washington\\
\href{http://yonatanbisk.com/piqa}{\texttt{http://yonatanbisk.com/piqa}}
}

\begin{document}

\maketitle

\begin{abstract}
To apply eyeshadow without a brush, should I use a \textit{cotton swab or a toothpick}? Questions requiring this kind of  \textbf{physical commonsense} pose a challenge to today's natural language understanding systems. While recent pretrained models (such as BERT) have made progress on question answering over more \emph{abstract} domains -- such as news articles and encyclopedia entries, where text is plentiful -- in more \emph{physical} domains, text is inherently limited due to reporting bias. 
Can AI systems learn to reliably answer physical commonsense questions without experiencing the physical world?

In this paper, we introduce the task of physical commonsense reasoning and a corresponding benchmark dataset \datasetlong{} or \dataset{} \pig{}. Though humans find the dataset easy (95\% accuracy), large pretrained models struggle ($\sim$77\%). We provide analysis about the dimensions of knowledge that existing models lack, which offers significant opportunities for future research.
\end{abstract}

\section{Introduction}
Before children learn language, they already start forming categories and concepts based on the physical properties of objects around them \cite{hespos2004}.  This model of the world grows richer as they learn to speak, but already captures \emph{physical commonsense} knowledge about everyday objects:
their physical properties, affordances, and how they can be manipulated. This knowledge is critical for day-to-day human life, including tasks such as problem solving (what can I use as a pillow when camping?) and expressing needs and desires (bring me a hard\textit{er} pillow).  Likewise, we hypothesize that modeling physical commonsense knowledge is a major challenge on the road to true AI-completeness, including robots that interact with the world and understand natural language.

Much of physical commonsense can be expressed in language, as the versatility of everyday objects and common concepts eludes other label schemes. However, due to issues of reporting bias, these commonsense properties - facts like `it is a bad idea to apply eyeshadow with a toothpick' are rarely directly reported. Although much recent progress has been made in Natural Language Processing through a shift towards large-scale pretrained representations from unlabeled text \cite{Radford2018,Devlin2018,Liu2019}, the bulk of the success of this paradigm has been on core \emph{abstract} tasks and domains. State-of-the-art models can reliably answer questions given an encyclopedia article \cite{rajpurkar2016squad} or recognize named entities  \cite{tjong-kim-sang-de-meulder-2003-introduction}, but it is not clear whether they can robustly answer questions that require physical commonsense knowledge.


\begin{figure}[t]
    \centering
    \includegraphics[width=\linewidth]{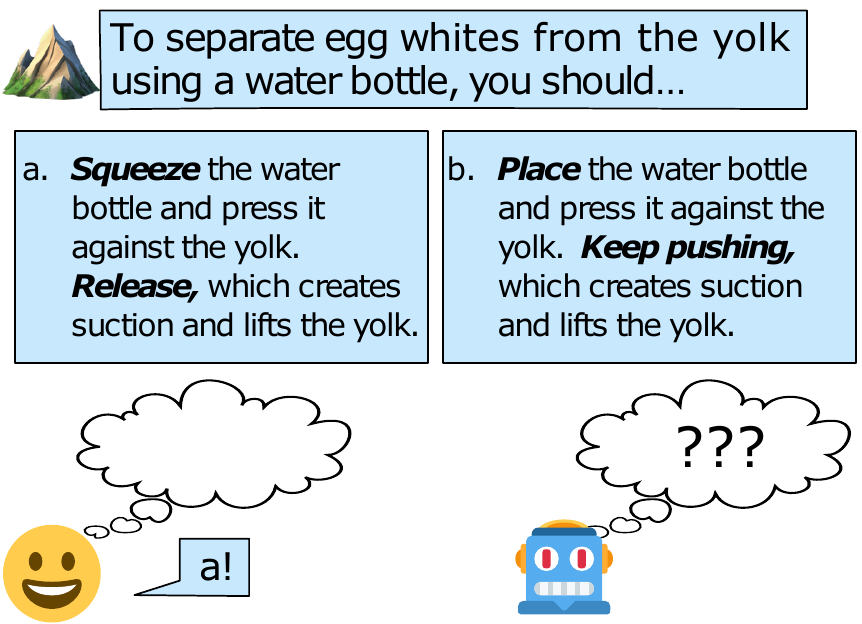}
    \caption{\dataset~\pig: Given a physical \textbf{goal} expressed in natural language, like `to separate egg whites...,' a model must choose the most sensible \textbf{solution}. Our dataset tests the ability of natural language understanding models to link text to a robust intuitive-physics model of the world. Here, humans easily pick answer \textbf{a)} because separating the egg requires \emph{pulling} the yolk out, while machines are easily fooled.}
    \label{tab:my_label}
\end{figure}

\begin{figure*}[t]
\centering
\begin{minipage}{.48\textwidth}
  \centering\small\fbox{
\begin{tabular}{@{}l@{\hspace{4pt}}p{22em}@{\hspace{4pt}}r@{}}
\multicolumn{2}{c}{\textbf{a. Shape, Material, and Purpose}}\\
\midrule
\goal & Make an outdoor pillow\\
\sol{1}  & Blow into a \textcolor{red}{tin can} and tie with rubber band & \rex{}\\
\sol{2}  & Blow into a \textcolor{blue}{trash bag} and tie with rubber band & \gcheck{}\\
             & \\
\goal    & To make a hard shelled taco,\\
\sol{1}  &  put seasoned beef, cheese, and lettuce \textcolor{red}{onto} the hard shell.  & \rex{}\\
\sol{2}  &  put seasoned beef, cheese, and lettuce \textcolor{blue}{into} the hard shell. & \gcheck{}\\
             & \\
\goal & How do I find something I lost on the carpet?\\
\sol{1}  &  Put a \textcolor{red}{solid seal} on the end of your vacuum and turn it on. & \rex{}\\
\sol{2}  &  Put a \textcolor{blue}{hair net} on the end of your vacuum and turn it on.  & \gcheck{}\\
\end{tabular}}
\end{minipage}%
\hspace{.039\textwidth}\begin{minipage}{.48\textwidth}
  \centering\small\fbox{
\begin{tabular}{@{}l@{\hspace{4pt}}p{22em}@{\hspace{4pt}}r@{}}
\multicolumn{2}{c}{\textbf{b. Commonsense Convenience}}\\
\midrule
\goal{} &How to make sure all the clocks in the house are set accurately? \\ \\
\sol{1}  & Get a solar clock for a reference and place it just outside a window that gets lots of sun. Use a system of call and response once a month, having one person stationed at the solar clock who yells out the correct time and have another person move to each of the indoor clocks to check if they are showing the right time. Adjust as necessary. &\rex{} \\ 
\sol{2}  &  Replace all wind-ups with digital clocks. That way, you set them once, and that's it. Check the batteries once a year or if you notice anything looks a little off. & \gcheck{}\\
             & \\
\end{tabular}
}
\end{minipage}
\caption{\dataset{} covers a broad array of phenomena.  Above are two categories of example QA pairs. \textbf{Left} are examples that require knowledge of basic properties of the objects (flexibility, curvature, and being porous), while on the \textbf{Right} both answers may be technically correct but one is more convenient and preferable. 
}
\label{fig:dataexamples}
\end{figure*}

To study this question and begin bridging the representational gap, we introduce \datasetlong{}, or \dataset{} \pig{} to evaluate language representations on their knowledge of physical commonsense.  We focus on everyday situations with a preference for atypical solutions. 
Our dataset is inspired by instructables.com,  which provides users with instructions on how to build, craft, bake, or manipulate objects using everyday materials.  We asked annotators to provide semantic perturbations or alternative approaches which are otherwise syntactically and topically similar to ensure physical knowledge is targeted.  The dataset is further cleaned of basic artifacts using the \texttt{AFLite} algorithm introduced in \cite{Sakaguchi2019WINOGRANDEAA,Sap2019} which is an improvement on adversarial filtering \cite{Zellers2018,Zellers2019b}. 

Throughout this work we first detail the construction of our new benchmark for physical commonsense. Second, we show that popular approaches to large-scale language pretraining, while highly successful on many \emph{abstract} tasks, fall short when a physical model of the world is required. Finally, our goal is to elicit further research into building language representations that capture details of the real world.  To these ends, we perform error and corpora analyses to provide insights for future work.

\section{Dataset}
We introduce a new dataset, \dataset{} \pig{}, for benchmarking progress in physical commonsense understanding. The underlying task is multiple choice question answering: given a question $\boldsymbol{q}$ and two possible solutions $\boldsymbol{s_1}, \boldsymbol{s_2}$, a model or a human must choose the most appropriate solution, of which exactly one is correct.
We collect data with how-to instructions as a scaffold, and use state-of-the-art approaches for handling spurious biases, which we will discuss below.

\subsection{Instructables as a source of physical commonsense}
Our goal is to construct a resource that requires concrete physical reasoning. To achieve this, we provide a prompt to the annotators derived from instructables.com.
The instructables website is a crowdsourced collection of instructions for doing everything from cooking to car repair.  In most cases, users provide images or videos detailing each step and a list of tools that will be required.  Most goals are simultaneously rare and unsurprising.  While an annotator is unlikely to have built a UV-Flourescent steampunk lamp or made a backpack out of duct tape, it is not surprising that someone interested in home crafting would create these, nor will the tools and materials be unfamiliar to the average person.  
Using these examples as the seed for their annotation, helps remind annotators about the less prototypical uses of everyday objects.
Second, and equally important, is that instructions build on one another.  This means that any QA pair inspired by an instructable is more likely to explicitly state assumptions about what preconditions need to be met to start the task and what postconditions define success.

\subsection{Collecting data through goal-solution pairs}
Unlike traditional QA tasks, we define our dataset in terms of Goal and Solution pairs (see Figure \ref{fig:dataexamples} for example Goal-Solution pairs and types of physical reasoning).  The Goal in most cases can be viewed as indicating a post-condition and the solutions indicate the procedure for accomplishing this.  The more detailed the goal, the easier it is for annotators to write both correct and incorrect solutions.  As noted above, the second component of our annotation design is reminding people to think creatively.  
We initially experimented with asking annotators for (task, tool) pairs via unconstrained prompts, but found that reporting bias swamped the dataset.  In particular, when thinking about how to achieve a goal, people most often are drawn to prototypical solutions and look for tools in the kitchen (e.g. forks and knives) or the garage (e.g. hammers and drills).  They rarely considered the literal hundreds of other everyday objects that might be in their own homes (e.g. sidewalk chalk, shower curtains, etc).  

\begin{figure}
    \centering
    \includegraphics[width=\linewidth]{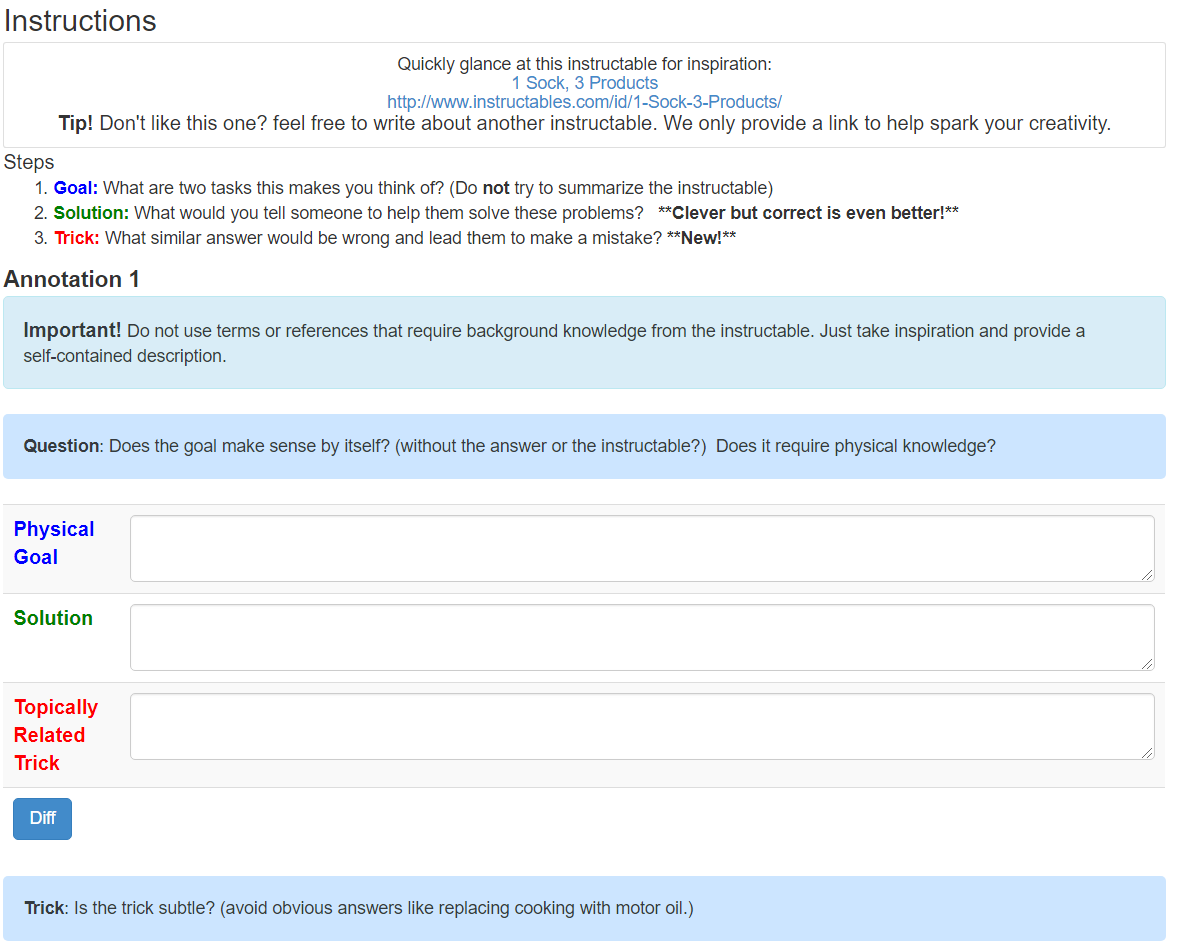}
    \caption{In the HIT design the instructable provides inspiration to think out-of-the-box (\textit{1 Sock, 3 Products}) and annotators are asked for 1. a \textit{physical} goal, 2. a valid solution, and 3. a trick.  The trick should sound reasonable, but be wrong often due to a subtle misunderstanding of preconditions or physics. Additional HITs (not shown) were run for qualification prior to this stage and validation afterwards.\footnotemark}
    \label{fig:HIT}
\end{figure}

To address this, and flatten the distribution of referenced objects (see Figure \ref{fig:topTags}), we prompt the annotations with links to instructables.  Specifically, annotators were asked to glance at the instructions of an instructable and pull out or have it inspire them to construct two component tasks.  They would then articulate the goal (often centered on atypical materials) and how to achieve it.  In addition, we asked them to provide a permutation to their own solution which makes it invalid, often subtly (Figure \ref{fig:HIT}). To further assist diversity we seed annotators with instructables drawn from six categories (costume, outside, craft, home, food, and workshop).  We asked that two examples be drawn per instructable to encourage one of them to come later in the process and require precise articulation of pre-conditions.

\footnotetext{In addition to this design, we also include a qualification HIT which contained well constructed and underspecified (goal, solution) pairs.  Annotators had to successfully ($>$80\%) identify which were well formed to participate in the main HIT.  Data was collected in batches of several thousand triples and validated by other annotators for correctness.  Users will low agreement were de-qualed.}
During validation, examples with low agreement were removed from the data. This often meant that correct examples were removed that required expert level knowledge of a domain (e.g. special woodworking terminology) which should not fall under the umbrella of ``commonsense." Because, we focus on human generated tricks, annotators were free to come up with clever ways to hide deception.  Often, this meant making very subtle changes to the solution to render it incorrect.  In these cases, the two solutions may differ by as little as one word. We found that annotations used both simple linguistic tricks (e.g. negation and numerical changes) and often swapped a key action or item for another that was topically similar but not helpful for completing the given goal. For this reason, our interface also includes a \texttt{diff} button which highlights where the solutions differ.  This improved annotator accuracy and speed substantially.  Annotator pay averaged $>\!15$\$/hr according to both self-reporting on \url{turkerview.com} and our timing calculations.

\subsection{Statistics}
In total our dataset is comprised of over 16,000 training QA pairs with an additional $\sim$2K and $\sim$3k held out for development and testing, respectively.  Our goals, as tokenized by Spacy,\footnote{\url{https://spacy.io} --  all data was collected in English.} average 7.8 words and both correct and incorrect solutions average 21.3 words.  In total, this leads to over 3.7 million lexical tokens in the training data. 

\begin{figure}
\centering
\includegraphics[width=\linewidth]{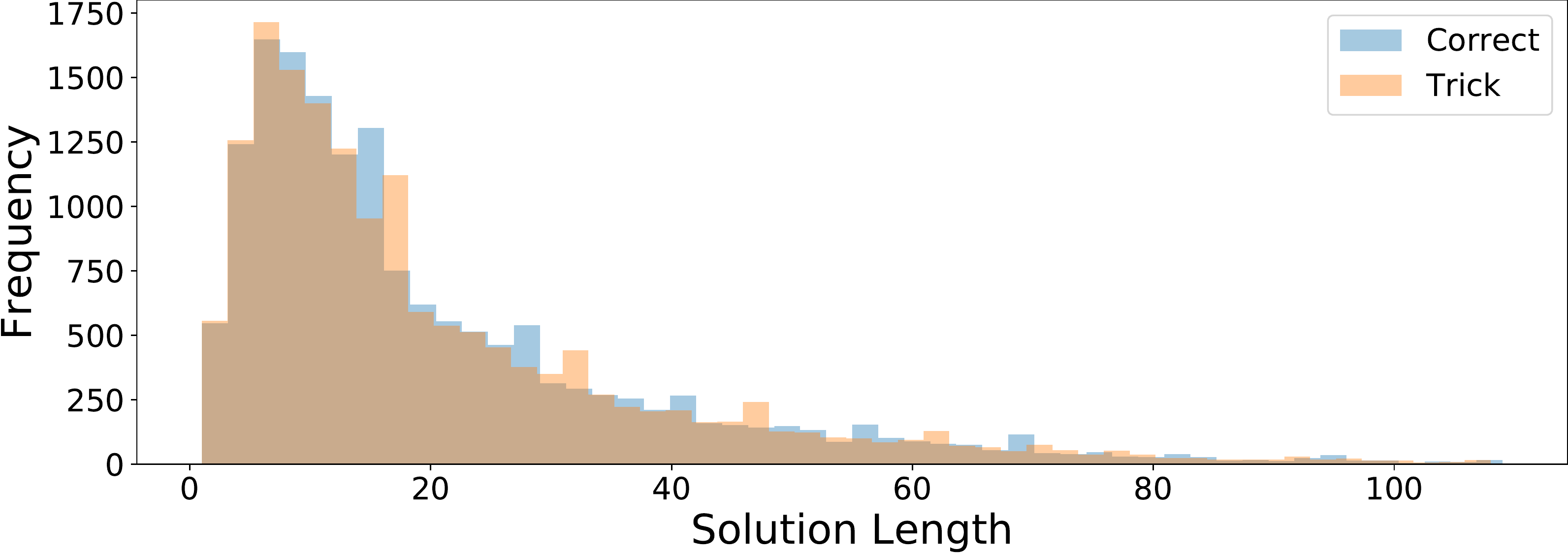}
\caption{Sentence length distributions for both correct solutions and tricks are nearly identical across the training set.}
\label{fig:sol_len}
\end{figure}

Figure \ref{fig:sol_len} shows a plot of the correct and incorrect sequence lengths (as tokenized by the GPT BPE tokenizer), with the longest 1\% of the data removed.  While there are minor differences, the two distributions are nearly identical.

\begin{figure*}[t]
\centering
\includegraphics[width=\textwidth]{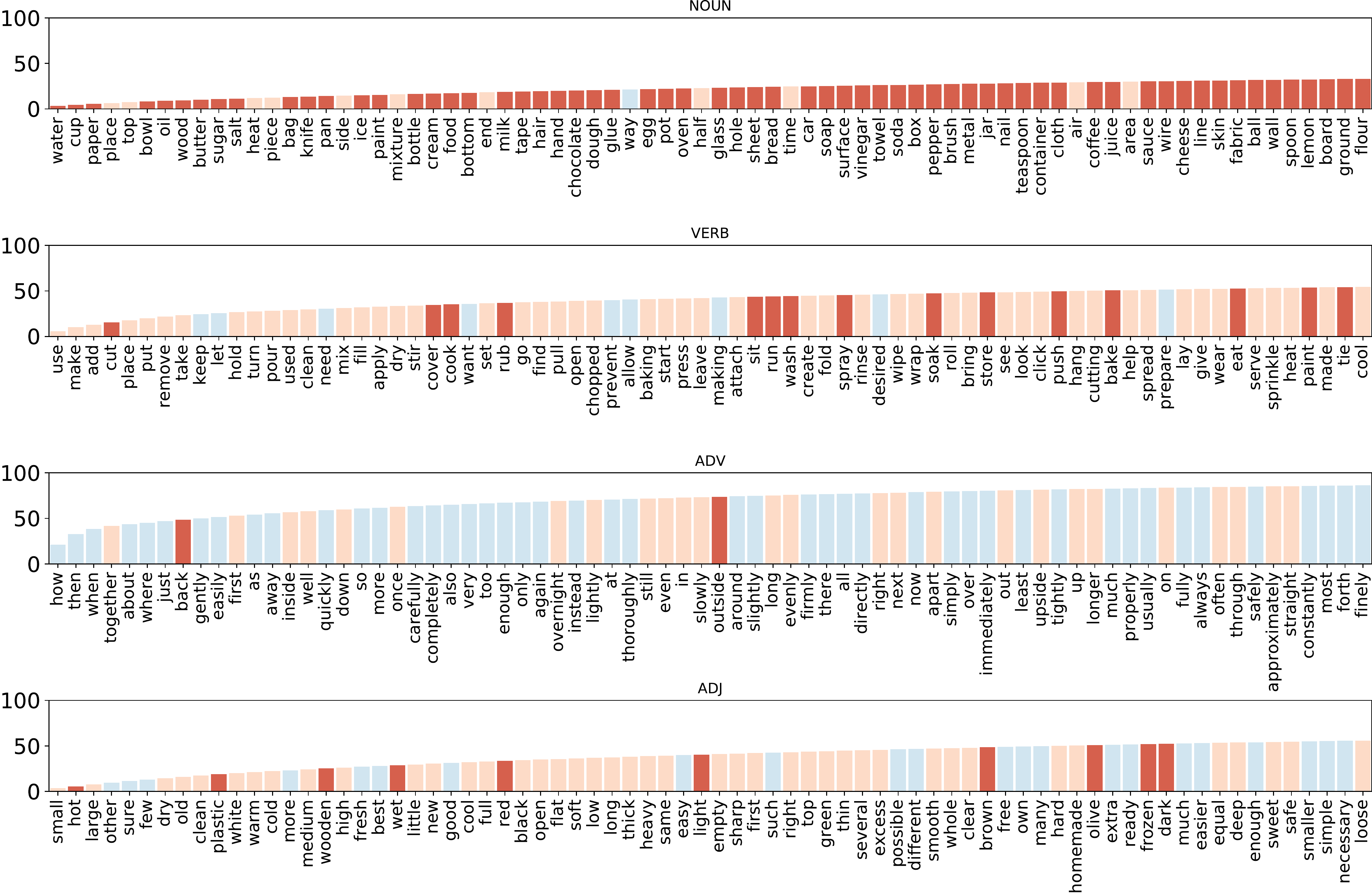}
\caption{Here we show the frequency distributions for the top seventy-five words tagged by Spacy as noun, verb, adverb or adjective.  We see that the vast majority of concepts focus on physical properties (\textit{e.g. small, hot, plastic, wooden}) and how objects can be manipulated (\textit{e.g. cut, cover, soak, push}).  Additionally, we see strongly zipfian behavior in all tags but the adverbs.  Words are colored by the average concreteness scores presented by \cite{concrete}.
}
\label{fig:topTags}
\end{figure*}

We also analyzed the overlap in the vocabulary and find that in all cases (noun, verb, adjective, and adverb) we see at least an 85\% overlap between words used in correct and incorrect solutions.  In total we have 6,881 unique nouns, 2,493 verbs, 2,263 adjectives, and 604 adverbs in the training data..  The most common of each are plotted in Figure \ref{fig:topTags} alongside their cumulative distributions. Again, this helps verify that the dataset revolves very heavily around physical phenomena, properties, and manipulations.  For example, the top adjectives include state (\textit{dry, clean, hot}) and shape (\textit{small, sharp, flat}); adverbs include temporal conditions (\textit{then, when}) and manner (\textit{quickly, carefully, completely}). These properties often differentiate correct from incorrect answers, as shown in examples throughout the paper. We also color words according to their concreteness score \cite{concrete}, though many ``abstract" words have  concrete realizations in our dataset.

\subsection{Removing Annotation Artifacts}

As noted previously, we use \texttt{AFLite} \cite{Sakaguchi2019WINOGRANDEAA} to remove stylistic artifacts and trivial examples from the data, which have been shown to artificially inflate model performance on previous NLI benchmarks \cite{hypothesis-only-baselines-in-natural-language-inference,gururangan-etal-2018-annotation}.  
The \texttt{AFLite} algorithm performs a systematic data bias reduction: it discards instances whose given feature representations are collectively highly indicative of the target label. In practice, we use 5,000 examples from the original dataset to fine-tune BERT-Large for this task and compute the corresponding embeddings of all remaining instances. \texttt{AFLite} uses an ensemble of linear classifiers trained on random subsets of the data to determine whether these pre-computed embeddings are strong indicators of the correct answer option. Instead of having to specifically identify the possible sources of biases, this approach enables unsupervised data bias reduction by relying on state-of-the-art methods to uncover undesirable annotation artifacts. For more information about \texttt{AFLite}, please refer to \cite{Sakaguchi2019WINOGRANDEAA}.

\section{Experiments}
In this section, we test the performance of state-of-the-art natural language understanding models on our dataset, \dataset. In particular, we consider the following three large-scale pretrained transformer models:

\begin{enumerate}[wide, labelwidth=!,labelindent=0pt,noitemsep,topsep=0pt,label=\textbf{\alph*}.]
    \item {\bf GPT} \cite{Radford2018} is a model that processes text left-to-right, and was pretrained using a language modeling objective. We use the original 124M parameter GPT model.
    \item {\bf BERT} \cite{Devlin2018} is a model that process text bidirectionally, and thus was pretrained using a special masked language modeling objective. We use BERT-Large with 340M parameters.
    \item {\bf RoBERTa} \cite{Liu2019} is a version of the BERT model that was made to be significantly more robust through pretraining on more data and careful validation of the pretraining hyperparameters. We use RoBERTa-Large, which has 355M parameters.
\end{enumerate}

We follow standard best practices in adapting these models for two-way classification. We consider the two solution choices independently: for each choice, the model is provided the goal, the solution choice, and a special \texttt{[CLS]} token. At the final layer of the transformer, we extract the hidden states corresponding to the positions of each \texttt{[CLS]} token. We apply a linear transformation to each hidden state and apply a softmax over the two options: this approximates the probability that the correct solution is option A or B. During finetuning, we train the model using a cross-entropy loss over the two options. For GPT, we follow the original implementation and include an additional language modeling loss, which improved training stability.

Generally, we found that finetuning was often unstable with some hyperparameter configurations leading to validation performance around chance, particularly for BERT. We follow best practices in using a grid search over learning rates, batch sizes, and the number of training epochs for each model, and report the best-scoring configuration as was found on the validation set. For all models and experiments, we used the \texttt{transformers} 
library and truncated examples at 150 tokens, which affects 1\% of the data.

Manual inspection of the development errors show that some ``mistakes" are actually correct but required a web-search to verify. Human performance was calculated by a majority vote. Annotators were chosen to participate that achieved $\geq$90\% on the qualification HIT from before. It is therefore, completely reasonable that automated methods trained on large web crawls may eventually surpass human performance here.
Human evaluation was performed on development data, and the train, development, and test folds were automatically produced by \texttt{AFLite}. 

\subsection{Results}
\begin{table}
\centering
\begin{tabular}{lccc}
\toprule
                 &         & \multicolumn{2}{c}{Accuracy (\%)} \\
Model            & Size    & Validation & Test \\
\midrule
Random Chance & & 50.0 & 50.0 \\
Majority Class & & 50.5 & 50.4 \\
\midrule
OpenAI GPT       & 124M   & 70.9  & 69.2 \\
Google BERT      & 340M   & 67.1  & 66.8 \\
FAIR RoBERTa & 355M   & 79.2  & 77.1 \\
\midrule
Human &  & \multicolumn{2}{c}{94.9} \\
\bottomrule
\end{tabular}
\caption{Results of state-of-the-art natural language understanding models on \dataset, compared with human performance. The results show a significant gap between model and human performance, of roughly 20 absolute points.}
\label{tab:results}
\end{table}

We present our results in Table~\ref{tab:results}.  As the dataset was constructed to be adversarial to BERT, it is not surprising that it performs the worst of three models despite generally outperforming GPT on most other benchmarks.  Comparing GPT and RoBERTa we see that despite more training data, a larger vocabulary, twice the number of parameters and careful construction of robust training, there is only a 8pt performance gain and RoBERTa still falls roughly 18 points short of human performance on this task.  As noted throughout, exploring this gap is precisely the purpose for \dataset{} existing and which facets of the dataset fool RoBERTa is the focus of the remainder of this paper. 

\section{Analysis}
\begin{figure}[t]
    \centering
    \includegraphics[width=\linewidth]{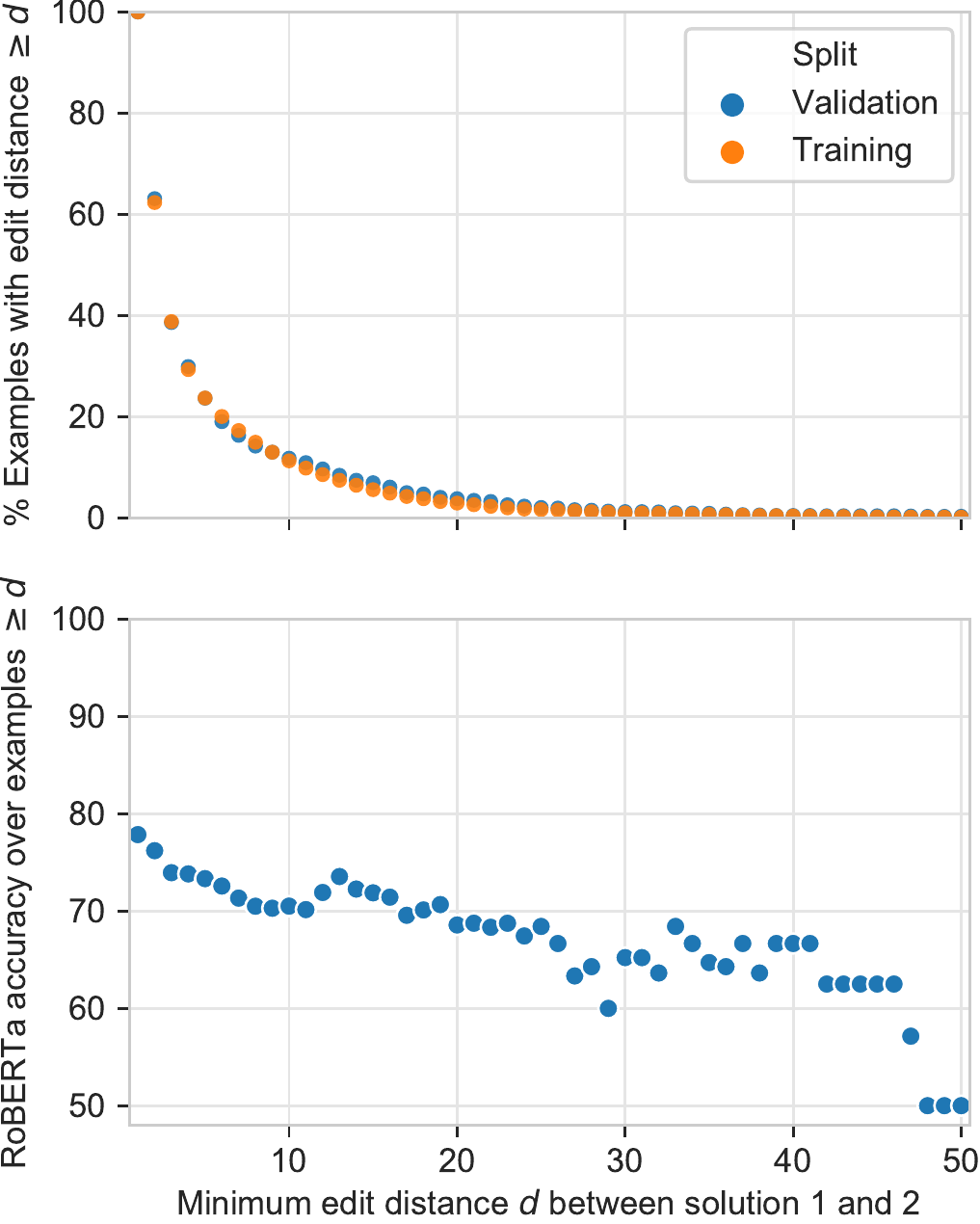}
    \caption{Breaking down \dataset~by edit distance between solution choices. \textbf{Top}: Cumulative histogram of examples in the validation and training sets, in terms of minimum edit distance $d$ between the two solution choices. The majority of the dataset consists of small tweaks between the two solution pairs; nevertheless, this is enough to confuse state-of-the-art NLP models. \textbf{Bottom}: RoBERTa accuracy over validation examples with a minimum edit distance of $d$. Dataset difficulty increases somewhat as the two solution pairs are allowed to drift further apart.}
    \label{fig:accuracybyeditdistance}
\end{figure}
In this section, we unpack the results of state-of-the-art models on \dataset. In particular, we take a look at the errors made by the top-performing model RoBERTa, as a view towards the physical commonsense knowledge that can be learned through language alone.

\subsection{\dataset~as a diagnostic for physical understanding}
The setup of \dataset~allows us to use it to probe the inner workings of deep pretrained language models, and to determine the extent of their physical knowledge. In this way, our dataset can augment prior work on studying to what extent models such as BERT understand syntax \cite{goldberg2019assessing}. However, while syntax is a well studied problem within linguistics, physical commonsense does not have as rich a literature to borrow from, making its dimensions challenging to pin down.

\textbf{Simple concepts.} Understanding the physical world requires a deep understanding of simple concepts, such as ``water'' or ``ketchup,'' and their affordances and interactions with respect to other concepts. 
Though our dataset covers \emph{interactions} between and with common objects, we can analyze the space of concepts in the dataset by performing a string alignment between solution pairs. Two solution choices that differ by editing a single phrase must by definition test the commonsense understanding of that phrase.

In Figure~\ref{fig:accuracybyeditdistance} we show the distribution of the edit distance between solution choices. We compute edit distance over tokenized and lowercased strings with punctuation removed. We use a cost of 1 for edits, insertions, and deletions. Most of the dataset covers simple edits between the two solution choices: roughly 60\% of the dataset in both validation and training involves a 1-2 word edit between solutions. In the bottom of Figure~\ref{fig:accuracybyeditdistance}, we show that the dataset complexity generally increases with the edit distance between the solution pairs. Nevertheless, the head of the distribution represents a space that is simple to study.

\begin{figure}[t]
    \centering
    \includegraphics[width=\linewidth]{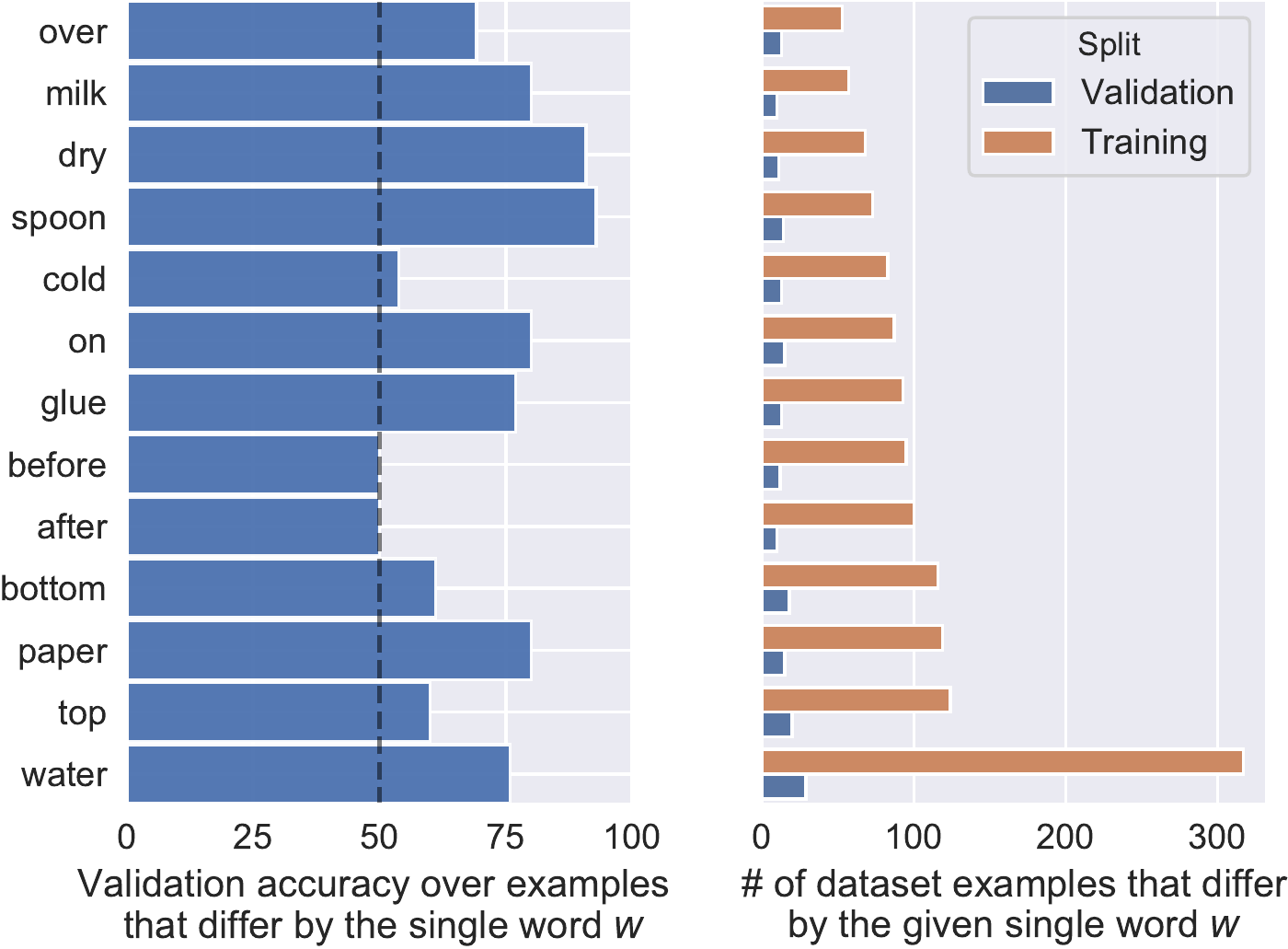}
    \caption{Common concepts as a window to RoBERTa's understanding of the physical world. We consider validation examples $(\boldsymbol{q}, \boldsymbol{s}_1, \boldsymbol{s}_2)$ wherein $\boldsymbol{s}_1$ and $\boldsymbol{s}_2$ differ from each other by a given word $w$. \textbf{Left}, we show the validation accuracy for common words $w$, while the number of dataset examples are shown \textbf{right.} Though certain concepts such as \emph{water} occur quite frequently, RoBERTa nevertheless finds those concepts difficult, with 75\% accuracy. Additionally, on common relations such as `cold', `on', `before', and `after' RoBERTa performs roughly at chance.}
    \label{fig:words}
\end{figure}

\textbf{Single-word edits.} In Figure~\ref{fig:words}, we plot the accuracy of RoBERTa among dataset examples that differ by a single word. More formally, we consider examples $(\boldsymbol{q}, \boldsymbol{s}_1, \boldsymbol{s}_2)$ whereby moving from $\boldsymbol{s}_1$ to $\boldsymbol{s}_2$, or vice versa, requires editing a given word $w$.\footnote{We additionally allow for an additional insertion; this helps to capture simple phrases like going from `water' to `olive oil.' Nevertheless, these multiword expressions tend to be less common, which is why we omit them in Figure~\ref{fig:words}.} We show examples of words $w$ that occur frequently in both the training and validation splits of the dataset, which allows RoBERTa to refine representations of these concepts during training and gives us a large enough sample size to reliably estimate model performance.

As shown, RoBERTa struggles to understand certain highly flexible relations. In particular, Figure~\ref{fig:words} highlights the difficulty of correctly answering questions that differ by the words `before,' `after', `top`, and `bottom': RoBERTa performs nearly at chance when encountering these.

Interestingly, the concepts shown in Figure~\ref{fig:words} suggest that RoBERTa also struggles to understand many common, more versatile, physical concepts. Though there are 300 training examples wherein the solution choices $\boldsymbol{s}_1, \boldsymbol{s}_2$ differ by the word `water.' RoBERTa performs worse than average on these replacements. On the other hand, RoBERTa does much better at certain nouns, such as `spoon.' 

\begin{figure}[t]
    \centering
    \includegraphics[width=\linewidth]{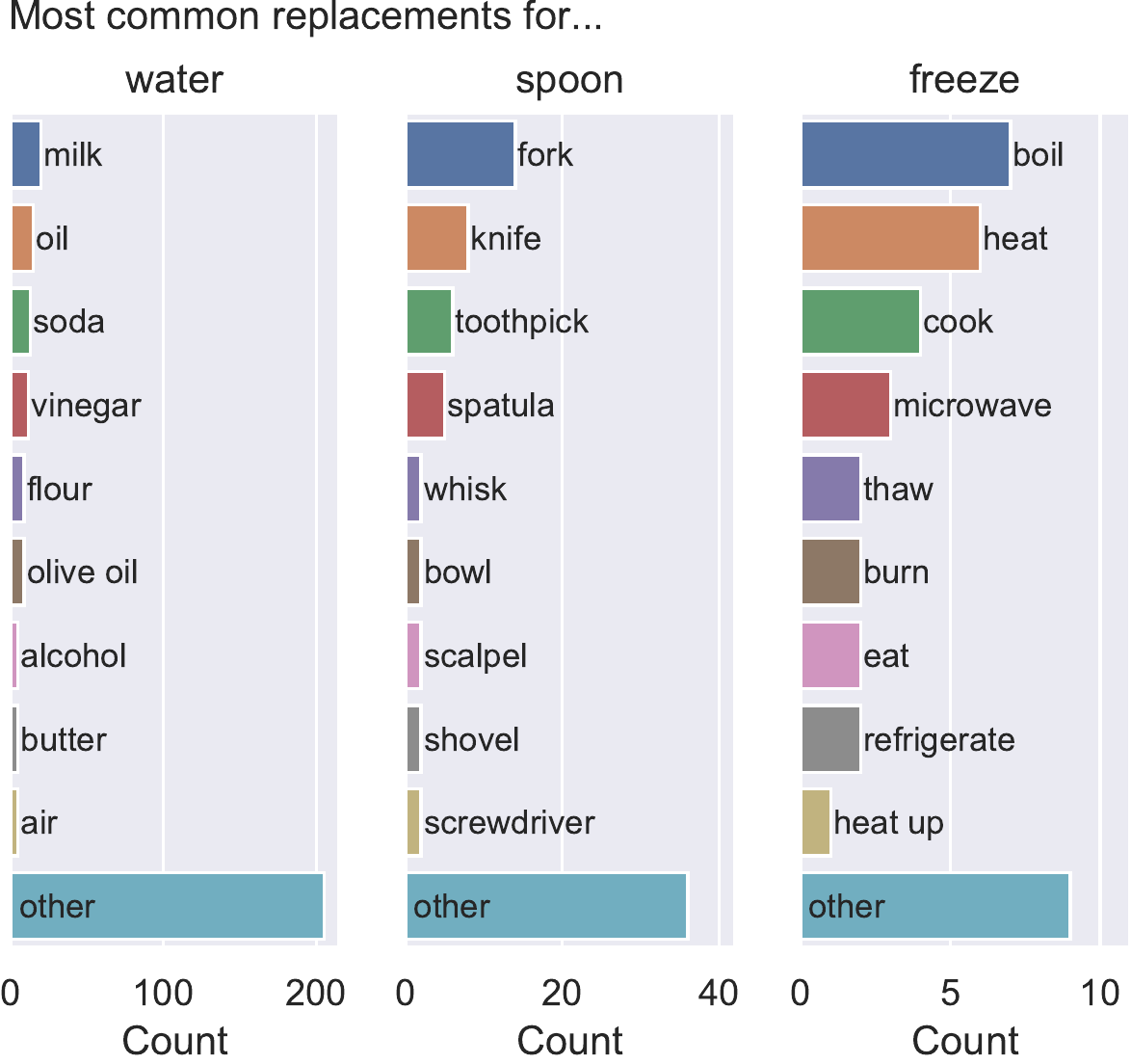}
    \caption{The most common replacements for three selected words: `water,' `spoon,' and `freeze.' These cover several key dimensions: `water' is a broad noun with many properties and affordances, whereas `spoons' are much narrower in scope. Perhaps as a result, RoBERTa performs much butter at examples where `spoon' is the pivot word (90\%) versus `water' (75\%). Freeze has an accuracy of 66\% on the validation set, and shows that verbs are challenging as well. }
    \label{fig:commonreplacements}
\end{figure}

\textbf{Common replacements in PIQA.} We dig into this further in Figure~\ref{fig:commonreplacements}, where we showcase the most common replacements for three examples: `water,' `spoon,' and `freeze.' While `water' is prevalent in the training set, it is also highly versatile. One can try to substitute it with a variety of different household items, such as `milk' or `alcohol,' often to disastrous effects. However, `spoons' have fewer challenging properties. A spoon cannot generally be substituted with a utensil that is sharp or has prongs, such as a fork, a knife, or a toothpick. RoBERTa obtains high accuracy on `spoon' examples, which suggests that it might understand this simple affordance, but does not capture the long tail of affordances associated with `water.'

\begin{figure*}[t]
\centering
\begin{minipage}{.48\textwidth}
  \centering\small
\begin{tabular}{@{}l@{\hspace{4pt}}p{23em}@{\hspace{4pt}}r@{}}
\multicolumn{2}{c}{\textbf{Correct examples}}\\
\midrule
\goal & Best way to pierce ears.\\
\sol{1}  & \correctans{It is best to go to a professional to get your ear pierced to avoid medical problems later.} & \gcheck{}\\
\sol{2}  & The best way to pierce your ears would be to insert a needle half inch thick into the spot you want pierced. & \rex{}\\
& \\
\goal & How do you reduce wear and tear on the nonstick finish of muffin pans?\\
\sol{1}  & \correctans{Make sure you use \textbf{\textit{paper liners}} to protect the nonstick finish when baking muffins and cupcakes in muffin pans.} & \gcheck{}\\
\sol{2}  & Make sure you use \textbf{\textit{grease and flour}} to protect the nonstick finish when baking muffins and cupcakes in muffin pans. & \rex{}\\
\end{tabular}
\end{minipage}%
\hspace{.039\textwidth}\begin{minipage}{.48\textwidth}
  \centering\small
\begin{tabular}{@{}l@{\hspace{4pt}}p{23em}@{\hspace{4pt}}r@{}}
\multicolumn{2}{c}{\textbf{Incorrect examples}}\\
\midrule
\goal & How can I quickly and easily remove strawberry stems?\\
\sol{1}  & \incans{Take a straw and from the \textbf{\textit{top}} of the strawberry push the straw through the center of the strawberry until the stem pops off.} & \rex{}\\
\sol{2}  & Take a straw and from the \textbf{\textit{bottom}} of the strawberry push the straw through the center of the strawberry until the stem pops off. & \gcheck{}\\
& \\ 
\goal & how to add feet to a coaster.\\
\sol{1}  & cut four slices from a glue stick, and attatch to the coaster with glue. & \gcheck{}\\
\sol{2}  & \incans{place a board under the coaster, and secure with zip ties and a glue gun.} & \rex{}\\
\end{tabular}
\end{minipage}
\caption{Qualitative analysis of RoBERTa's predictions with. \textbf{Left}: Two examples that RoBERTa gets right. \textbf{Right}: two examples that RoBERTa gets incorrect. Short phrases that differ between solution 1 and solution 2 are shown in \textbf{\text{bold and italics}}.
}
\label{fig:qualitativeanalysis}
\end{figure*}

\subsection{Qualitative results}
Our analysis thus far has been on simple-to-analyze single word expressions, where we have shown that the state-of-the-art language model, RoBERTa, struggles at a nuanced understanding of key commonsense concepts, such as relations. To further probe the knowledge gap of these strong models, we present qualitative examples in Figure~\ref{fig:qualitativeanalysis}. The examples are broadly representative of larger patterns: RoBERTa can recognize clearly ridiculous generations (Figure~\ref{fig:qualitativeanalysis}, top left) and understands differences between some commonsense concepts (bottom left). It's important to note, that in both cases the correct answer is prototypical and something we might expect the models to have seen before.

However, it struggles to tell the difference between subtle relations such as top and bottom (top right of Figure~\ref{fig:qualitativeanalysis}). Moreover, it struggles with identifying non-prototypical situations (bottom right). Though using a gluestick as feet for a coaster is uncommon, to a human familiar with these concepts we can visualize the action and its result to verify that the goal has been achieved.
Overall, these examples suggest that physical understanding -- particularly involving novel combinations of common objects -- challenges models that were pretrained on text only.

\section{Related Work}
Physical understanding is broad domain that touches on everything from scientific knowledge \cite{Schoenick2016} to the interactive acquisition of knowledge by embodied agents \cite{Thomason2016}. To this end, work related to the goals of our benchmark span the NLP, Computer Vision and Robotics communities.

\textbf{Language.} Within NLP, in addition to large scale models, there has also been progress on reasoning about cause and effect effects/implications within these models \cite{Bosselut2019}, extracting knowledge from them \cite{Petroni2019}, and investigating where large scale language models fail to capture knowledge of tools and elided procedural knowledge in recipes \cite{Bisk2019}.  The notion of procedural knowledge and instruction following is a more general related task within vision and robotics.  From text alone, work has shown that much can be understood about the implied physical situations of verb usage \cite{Forbes2017} and relative sizes of objects \cite{Elazar2019}.

\textbf{Vision.} Physical knowledge can be discovered and evaluated within the visual world. Research has studied predicting visual relationships in images \cite{krishnavisualgenome} and as well as actions and their dependent objects \cite{yatskar2016}. Relatedly, the recent HAKE dataset \cite{li2019hake} specifically annotates which object/body-parts are essential to completing or defining an action.  
Image data also allows for studying the concreteness of nouns and provides a natural path forward for further investigation \cite{hessel-etal-2018-quantifying}.
Related to physical commonsense, research in \emph{visual commonsense} has studied intuitive physics \cite{NIPS2017_6620}, cause-effect relationships \cite{Mottaghi2016}, and what can be reasonably inferred beyond a single image \cite{Zellers2019a}. 

\textbf{Robotics.} Learning from interaction and intuitive physics \cite{Agrawal2016} can also be encoded as priors when exploring the world \cite{Byravan2018} and internal models of physics, shape, and material strength enable advances in tool usage \cite{Toussaint2018} or construction \cite{Nair2019}. Key to our research aims in this work is helping to build language tools which capture enough physical knowledge to speed up the bootstrapping of robotic-language applications.  Language tools should provide strong initial priors for learning \cite{Tellex2011,Matuszek2018} that are then refined through interaction and dialogue \cite{Gao2016}.

\section{Conclusion}
We have evaluated against large-scale pretrained models as they are in vogue as the de facto standard of progress within NLP, but are primarily interested in their performance and failings as a mechanism for advancing the position that learning about the world from language alone, is limiting.  Future research, may ``match" humans on our dataset by finding a large source of in-domain data and fine-tuning heavily, but this is very much \textit{not the point}.  Philosophically, knowledge should be learned from interaction with the world to eventually be communicated with language.

In this work we introduce the \datasetlong{} or \dataset{} \pig{} benchmark for evaluating and studying physical commonsense understanding in natural language models.  We find the best available pretrained models lack an understanding of some of the most basic physical properties of the world around us.  Our goal with \dataset{} is to provide insight and a benchmark for progress towards language representations that capture knowledge traditionally only seen or experienced, to enable the construction of language models useful beyond the NLP community.

\section{Acknowledgements}
We thank the anonymous reviewers for their insightful suggestions.  
This research was supported in part by NSF (IIS-1524371, IIS-1714566), DARPA under the CwC program through the ARO (W911NF-15-1-0543), DARPA under the MCS program through NIWC Pacific (N66001-19-2-4031), and the NSF-GRFP No. DGE-1256082.
Computations on \url{beaker.org} were supported in part by Google Cloud.

\bibliography{refs}
\bibliographystyle{aaai}

\end{document}